\documentclass[10pt, journal]{IEEEtran} % cspaper,compsoc

\usepackage{times}
\usepackage{epsfig}
\usepackage{graphicx}
\usepackage{amsmath}
\usepackage{amssymb}
\usepackage{hhline}
\usepackage{booktabs}
\usepackage{epstopdf}
\usepackage{subcaption}
\usepackage{caption}
\usepackage{color}
\usepackage{cite}
\usepackage{sidecap}
\usepackage{url}

\def\imagebox#1#2{\vtop to #1{\null\hbox{#2}\vfill}}

\begin{document}
%--------------------------------------------------------------------------
\title{A Discriminative Representation of Convolutional Features for Indoor Scene Recognition}

\author{S.~H.~Khan, %\IEEEmembership{Member,~IEEE,}
        M.~Hayat, %~\IEEEmembership{Fellow,~OSA,}
        M. Bennamoun, \IEEEmembership{Member,~IEEE,}
        R.~Togneri,
        and~F.~Sohel, \IEEEmembership{Senior Member,~IEEE}

\thanks{S.~H.~Khan, M.~Hayat, M.~Bennamoun and F.~Sohel are with the School of Computer Scince and Software Engineering (CSSE), University of Western Australia, Crawley, 6009. E-mail: \{salman.khan, munawar.hayat, mohammed.bennamoun, ferdous.sohel\}@uwa.edu.au}
\thanks{R. Togneri is with the School of Electrical, Electronics and Computer Engineering (EECE), University of Western Australia, Crawley, 6009. E-mail: roberto.togneri@uwa.edu.au}
\thanks{}} % ; revised --.Manuscript received January 29, 2015

%\markboth{Journal of \LaTeX\ Class Files,~Vol.~13, No.~9, September~2014}%
%{Shell \MakeLowercase{\textit{et al.}}: Bare Demo of IEEEtran.cls for Journals}

\maketitle

%--------------------------------------------------------------------------
\begin{abstract}
Indoor scene recognition is a multi-faceted and challenging problem due to the diverse intra-class variations and the confusing inter-class similarities.
This paper presents a novel approach which exploits rich mid-level convolutional features to categorize indoor scenes.
{\color{black} Traditionally used convolutional features preserve the global spatial structure, which is a desirable property for general object recognition.
However, we argue that this structuredness is not much helpful when we have large variations in scene layouts, e.g., in indoor scenes.}
We propose to transform the structured convolutional activations to another highly discriminative feature space.
The representation in the transformed space not only incorporates the discriminative aspects of the target dataset, but it also encodes the features in terms of the general object categories that are present in indoor scenes.
To this end, we introduce a new large-scale dataset of 1300 object categories which are commonly present in indoor scenes.
Our proposed approach achieves a significant performance boost over previous state of the art approaches on five major scene classification datasets.
\end{abstract}

\begin{IEEEkeywords}
Scene classification, convolutional neural networks, indoor objects dataset, feature representations, dictionary learning, sparse coding
\end{IEEEkeywords}
\IEEEpeerreviewmaketitle

%-------------------------------------------------------------------------
\section{Introduction}
%Scene Recognition Problem, its introduction and importance.

%high intra-class variability
%inter-class similarity

\begin{figure}[t]
\includegraphics[trim=49em 38em 9em 3em, clip, width=\columnwidth]{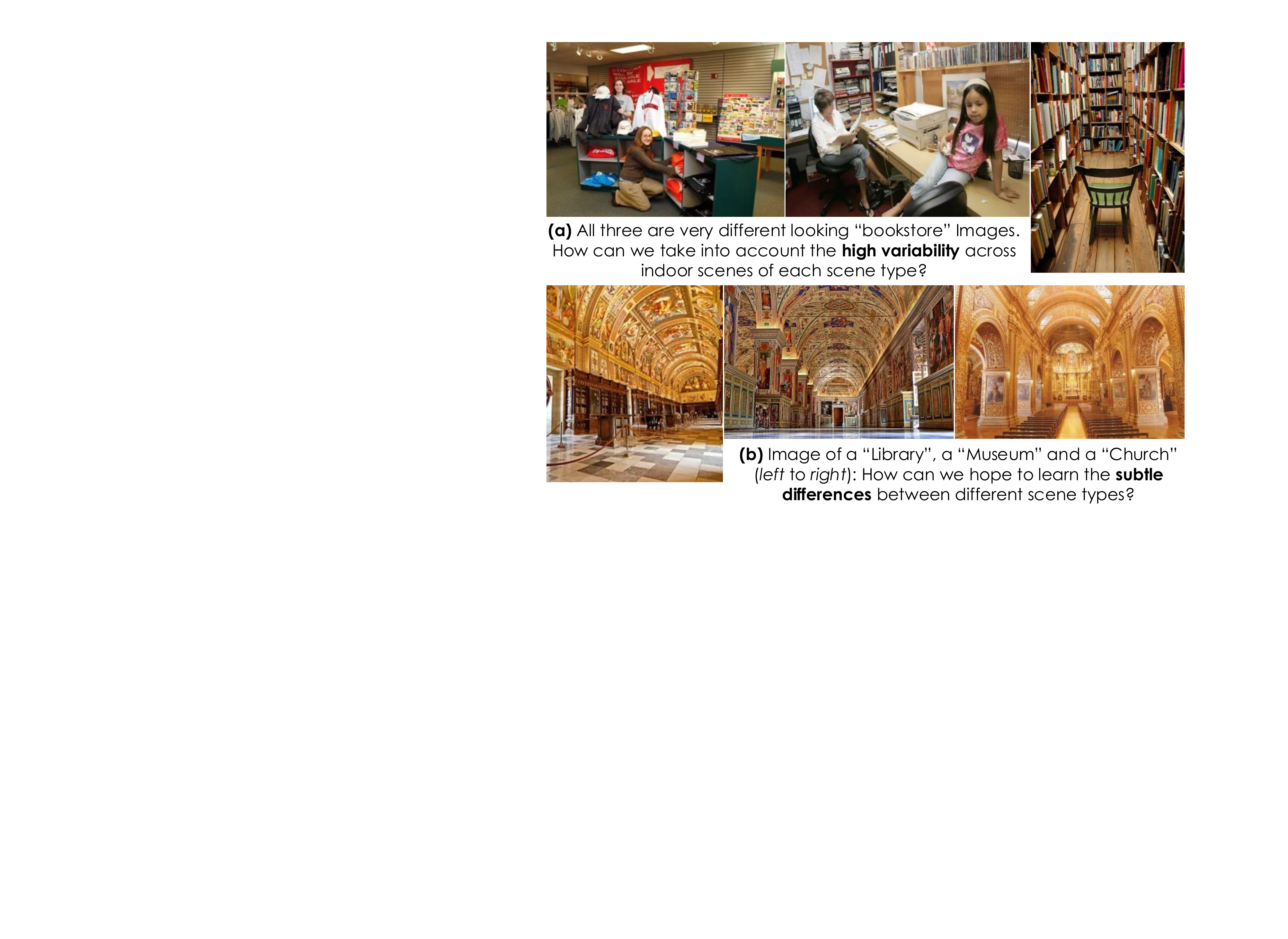}
%\vspace{-2em}
\caption{
\textbf{`Where am I located indoors?'}, we want to answer this question by assigning a semantic class label to a given color image.
An indoor scene categorization framework must take into account high intra-class variablity and should be able to tackle confusing inter-class similarities. This paper introduces a methodology to achieve these challenging requisites. (example images from MIT-67 dataset)}
\label{fig:var_inter_intra}
%\vspace{-1em}
\end{figure}

This paper proposes a novel method which captures the discriminative aspects of an indoor scene to correctly predict its semantic category (e.g.,  bedroom, kitchen etc.).
This categorization can greatly assist in context aware object and action recognition, object localization, robotic navigation and manipulation \cite{xiao2010sun, wu2011centrist}.
However, owing to the large variabilities between images of the same class and the confusing similarities between images of different classes, the automatic categorization of indoor scenes is a very challenging problem \cite{quattoni2009recognizing, xiao2010sun}.
Consider, for example, the images shown in Fig.~\ref{fig:var_inter_intra}.
 The images of the top row (Fig.~\ref{fig:var_inter_intra}~a) belong to the same class \emph{`bookstore'} and exhibit a large data variability in the form of object occlusions, cluttered regions, pose changes and varying appearances.
 The images in the bottom row (Fig.~\ref{fig:var_inter_intra}~b) are of three different classes and have large visual similarities.
A high performance classification system should therefore be able to cope with the inherently challenging nature of indoor scenes.

%-------------------------------------------------------------------------
%-------------------------------------------------------------------------

To deal with the challenges of indoor scenes, previous works \cite{fei2005bayesian, lazebnik2006beyond, oliva2001modeling, quattoni2009recognizing, margolin2014otc} propose to encode either local or global spatial and appearance information.
In this paper, we argue that neither of those two representations provide the best answer to effectively handle indoor scenes. The global representations are unable to model the subtle details, and the low-level local representations cannot capture object-to-object relations and the global structures \cite{quattoni2009recognizing, li2010object, singh2012unsupervised}.
We therefore
%propose to
devise mid-level representations that carry the necessary intermediate level of detail. These mid-level representations neither ignore the local cues nor lose the important scene structure and object category relationships.

 Our proposed mid-level representations are derived from densely and uniformly extracted image patches. In order to extract a rich feature representation from these patches, we use deep Convolutional Neural Networks (CNNs).
CNNs provide excellent generic mid-level feature representations and have recently shown great promise for large-scale classification and detection tasks \cite{oquab2013learning, Chatfield14, ILSVRCarxiv14, he2014spatial}.
They however tend to preserve the global spatial structure of the images \cite{zeiler2014visualizing}, which is not desirable when there are large intra-class variations e.g., in the case of indoor scene categorization (Fig.~\ref{fig:var_inter_intra}).
We therefore propose a method to discount this global spatial structure, while simultaneously retaining the intermediate scene structure which is necessary to model the mid-level scene elements.
For this purpose, we encode the extracted mid-level representations in terms of their association with codebooks\footnote{A codebook is a collection of distinctive mid-level patches. } of Scene Representative Patches (SRPs).
This enhances the robustness of the convolutional feature representations, while keeping intact their discriminative power.

It is interesting to note that some previous works hint towards the incorporation of `wide context' \cite{li2010object, kumar2005hierarchical, yao2012describing} for scene categorization. Such high-level context-aware reasoning has been shown to improve the classification performance.
However in this work, we show that for the case of highly variant indoor-scenes, mid-level context relationships prove to be the most decisive factor in classification.
The intermediate level of the scene details help in learning the subtle differences in the scene composition and its constituent objects.
In contrast, global structure patterns can confuse the learning/classification algorithm due to the high inter-class similarities (Sec.~\ref{subsec:results}).

As opposed to existing feature encoding schemes, we propose to form multiple codebooks of SRPs.
 We demonstrate that forming multiple smaller codebooks (instead of one large codebook) proves to be more efficient and produces a better performance (Sec. \ref{Ablative Analysis}).
Another key aspect of our feature encoding approach is the combination of supervised and unsupervised SRPs in our codebooks.
The unsupervised SRPs are collected from the training data itself, while the supervised SRPs are extracted from a newly introduced dataset of `Object Categories in Indoor Scenes' (OCIS).
The supervised SRPs provide semantically meaningful information, while the unsupervised SRPs relate more to the discriminative aspects of the different scenes that are present in the target dataset.
The efficacy of the proposed approach is demonstrated through extensive experiments on five challenging scene classification datasets. Our experimental results show that the proposed approach consistently achieves state of the art performance.

The \textbf{major contributions} of this paper are: \textbf{1).} We propose a new mid-level feature representation for indoor scene categorization using large-scale deep neural nets (Sec. \ref{sec:DUCA}),
\textbf{2)} Our feature description incorporates not only the discriminative patches of the target dataset but also the general object categories that are semantically meaningful (Sec. \ref{subsec:SRP}), % or human understandable
\textbf{3).} We collect the first large-scale dataset of object categories that are commonly present in indoor scenes. This dataset contains more than 1300 indoor object classes (Sec. \ref{subsubsec:OCISdatabase}),
\textbf{4).} To improve the efficiency and performance of our approach, we propose to generate multiple smaller codebooks and a feasible feature encoding (Sec. \ref{subsec:SRP}), and
\textbf{5).} We introduce a novel method to encode feature associations using max-margin hyper-planes (Sec. \ref{subsec:encoding_feat}).

%-------------------------------------------------------------------------
\section{Related Work}
\noindent
%\textbf{Indoor Scene Classification:}

Based upon the level of image description, existing scene classification techniques can be categorized into three types:
\textbf{1).} those which capture low level appearance cues, \textbf{2).} those which capture the high level spatial structure of the scene and \textbf{3).} those which capture mid-level relationships.
The techniques which capture low-level appearance cues \cite{fei2005bayesian, lazebnik2006beyond} perform poorly on the majority of indoor scene types since they fail to incorporate the high level spatial information.
The techniques which model the human perceptible global spatial envelope \cite{oliva2001modeling} also fail to cope with the high variability of indoor scenes.
The main reason for the low performance of these approaches is their neglect of the fine-grained objects, which are important for the task of scene classification.

Considering the need to extract global features as well as the characteristics of the constituent objects, Quattoni et al. \cite{quattoni2009recognizing} and Pandey et al. \cite{pandey2011scene} represented a scene as a combination of root nodes (which capture the global characteristics of the scene) and a set of regions of interest (which capture the local object characteristics).
However, the manual or automatic identification of these regions of interest makes their approach indirect and thus complicates the scene classification task.
Another example of indirect approach to scene recognition is the one proposed by Gupta et al. \cite{gupta2013perceptual}, where the grouping, segmentation and labeling outcomes are combined to recognize scenes.
Learned mid-level patches are employed for scene categorization by Juneja et al. \cite{juneja2013blocks},  Doersch et al. \cite{doersch2013mid} and Sun et al. \cite{sun2013learning}.
However these works involve a lot of effort in learning the distinctive primitives which includes a discriminative patch ranking and selection.
In contrast, our mid-level representation does not require any learning. Instead, we uniformly extract the mid-level patches densely from the images and show that these perform best when combined with supervised object representations.

{Deep Convolutional Neural Networks} have recently shown great promise in large-scale visual recognition and classification \cite{ouyang2014deepid, Chatfield14, ILSVRCarxiv14, krizhevsky2012imagenet, zeiler2011adaptive}.
%The recently used CNNs \cite{krizhevsky2012imagenet, zeiler2011adaptive} are an improved form of the Neocognitron model proposed by Fukushima \cite{fukushima1980neocognitron} and the ConvNet used for handwritten digits recognition by LeCun \cite{lecun1990handwritten}.
Although CNN features have demonstrated their discriminative power for images with one or multiple instances of the same object, they do preserve the spatial structure of the image, which is not desirable when dealing with the variability of indoor scenes \cite{he2014spatial}.
CNN architectures involve max-pooling operations to deal with the local spatial variability in the form of rotation and translation \cite{krizhevsky2012imagenet}.
However, these operations are not sufficient to cope with the large-scale deformations of objects and parts that are commonly present in indoor scenes \cite{he2014spatial, zeiler2011adaptive}.
In this work, we propose a novel representation which is robust to variations in the spatial structure of indoor scenes. It represents an image in terms of the association of its mid-level patches with the codebooks of the SRPs.

%\paragraph{Mid-level Representations}
%have demonstrated several successes in the conventional classification, detection and segmentation tasks.
%Popular frameworks include bag of features \cite{sivic2003video}, spatial pyramids \cite{lazebnik2006beyond}, deep belief nets \cite{hinton2006reducing}, convolutional and deconvolutional \cite{zeiler2011adaptive} networks.
%Attributes.

\begin{figure*}[ht]
\includegraphics[width = \textwidth]{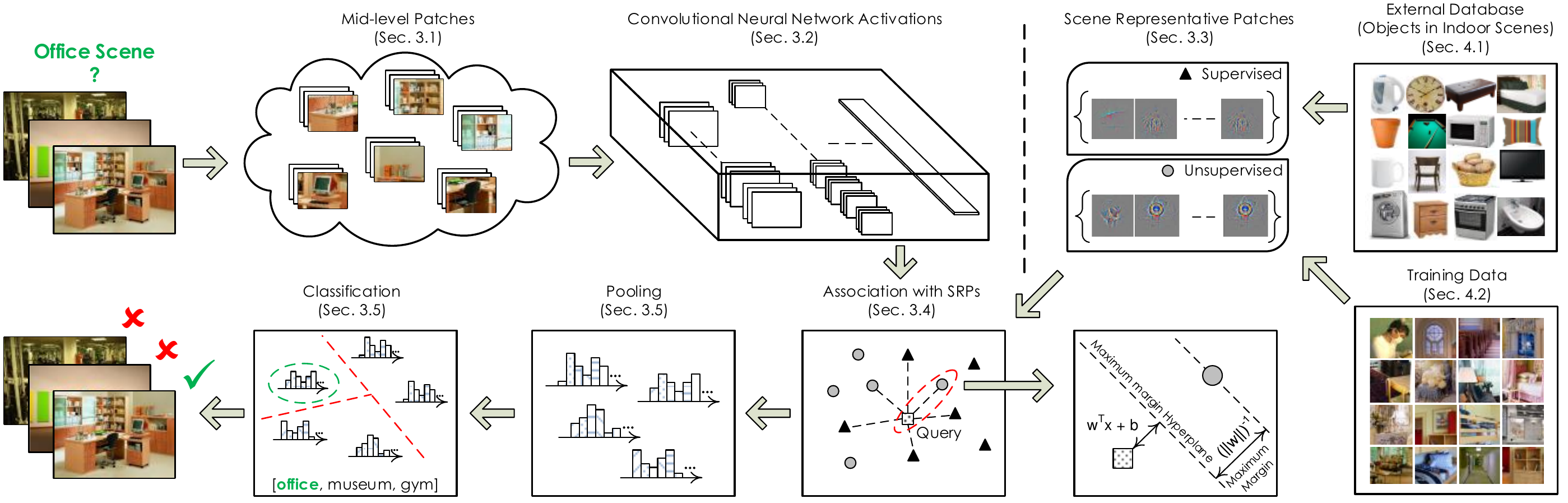}
%\vspace{-1em}
\caption{Deep Un-structured Convolutional Activations: Given an input image, we extract from it dense mid-level patches, represent the extracted patches by their convolutional activations and encode them in terms of their association with the codebooks of Scene Representative Patches (SRPs). The designed codebooks have both supervised and unsupervised SRPs. The resulting associations are then pooled and the class belonging decisions are predicted using a linear classifier.}
%\vspace{-1em}
\label{fig:overall_scheme}
\end{figure*}

%-------------------------------------------------------------------------
\section{The Proposed Method}%Deep Un-structured Convolutional Activations (DUCA)}
\label{sec:DUCA}

The block diagram of our proposed pipeline called `Deep Un-structured Convolutional Activations (DUCA)' is shown in Fig~\ref{fig:overall_scheme}. Our proposed method first densely and uniformly extracts mid-level patches (Sec~\ref{Dense Patch Extraction}), represents them by their convolutional activations (Sec~\ref{subsec:feat_extraction}) and then encodes them in terms of their association with the codebooks of SRPs (Sec~\ref{subsec:encoding_feat}), which are generated in supervised and unsupervised manners (Sec~\ref{subsec:SRP}). The detailed description of each component of the proposed pipeline is presented next.

%-------------------------------------------------------------------------
\subsection{Dense Patch Extraction}
\label{Dense Patch Extraction}

To deal with the high variability of indoor scenes, we propose to extract mid-level instead of global \cite{oliva2001modeling} or local \cite{fei2005bayesian, lazebnik2006beyond} feature representations.
Mid-level representations do not ignore object level relationships and the discriminative appearance
%and texture
based local cues (unlike the high level global descriptors), and do not ignore the holistic shape and scene structure information (unlike the low level local descriptors).
For each image, we extract dense mid-level patches using a sliding window of $224 \times 224$ pixels with a fixed step size of $32$. % or stride of 32
In order to extract a reasonable number of patches, the smaller dimension of the image is re-scaled to an appropriate length (700 pixels in our case).
Note that the idea of dense patch extraction is analogous to dense key-point extraction \cite{nowak2006sampling}, which has shown very promising performance over well-designed key-point extraction methods in a number of tasks (e.g., action recognition \cite{wang2009evaluation}).

%-----------------------------------------------------------------------------
%\subsection{Data Augmentation}
Before the dense patch extraction, we augment the images of the dataset with their flipped, cropped and rotated versions to the enhance generalization of our feature representation.
First, five cropped images (four from the corners and one from the center) of $\frac{2}{3}$ size are extracted from the original image.
Each original image is also subjected to CW and CCW rotations of $\frac{\pi}{6}$ radians and the resulting images are included in the augmented set.
The horizontally flipped versions of all  these eight images (1 original + 5 cropped + 2 rotated) are also included.
The proposed data augmentation results in a reasonable performance boost (see Sec. \ref{Ablative Analysis}).

\subsection{Convolutional Feature Representations}
\label{subsec:feat_extraction}
We need to map the raw image patches to a discriminative feature space where scene categories are easily separable. For this purpose, instead of using shallow or local feature representations, we use the convolutional activations from a trained deep CNN architecture. Learned representations based on CNNs have significantly outperformed hand-crafted representations in nearly all major computer vision tasks \cite{ Chatfield14, jia2014caffe}.
Our CNN architecture is similar to the `AlexNet' \cite{krizhevsky2012imagenet} (trained on ILSVRC 2012) and consists of 5 convolutional and 3 fully-connected layers.
The main difference compared to AlexNet is the dense connections between each pair of consecutive layers in the 8-layered network (in our case).
% The filters in the first layer are also a bit loyer i.e., 64 instead of 96
The densely and uniformly extracted patches from the images, are fed to the network's input layer after mean normalization.
The processed output from the network is taken from an intermediate fully connected layer ($7^{th}$ layer).
The resulting feature representation of each mid-level patch has a dimension of 4096.
% l2 normalization of CNN activations.

Although, CNN activations capture rich discriminative information, they are inherently highly structured.
The main reason is the sequence of operations involved in the hierarchical layers of CNN which preserve the global spatial structure of the image.
This constraining structure is a limitation when dealing with highly variable indoor scene images.
To address this, we propose to encode our patches (represented by their convolutional activations) to an alternate feature space which turns out to be even more discriminative (Sec.~\ref{subsec:encoding_feat}). Specifically, an image is encoded in terms of the association of its extracted patches with the codebooks of the Scene Representative Patches (SRPs).

%-------------------------------------------------------------------------
\subsection{Scene Representative Patches (SRPs)}
\label{subsec:SRP}
%An image of an indoor scene is characterized by a set of distinctive local parts.

An indoor scene is a collection of several distinct objects and concepts.
We are interested in extracting a set of image patches of these objects and concepts, which we call `Scene Representative Patches' (SRPs).
The SRPs can then be used as elements of a codebook to characterize any instance of an indoor scene.
Examples of these patches for a bedroom scene include a bed, wardrobe, sofa or a table.
Designing a comprehensive codebook of these patches is a very challenging task.
There can be two possible solutions: \textbf{first}, automatically learn to discover a number of discriminative patches from the training data and \textbf{second}, manually prepare an exhaustive vocabulary of all objects which can be present in indoor scenes.
These solutions are quite demanding.
First, because of the possibility of a very large number of objects, and second, this may require automatic object detection, localization or distinctive patch selection, which in itself is very challenging and computationally expensive.

In this work, we propose a novel approach to compile a comprehensive set of SRPs. Our proposed approach avoids the drawbacks of the above mentioned strategies and successfully combines their strengths i.e., it is computationally very efficient while being highly discriminative and semantically meaningful. Our set of SRPs has two main components, compiled in a \emph{supervised} and an \emph{unsupervised} manner.
These components are described next.

%--------------------------------------------------------------------
\subsubsection{Supervised SRPs}
\label{Supervised SRPs Extraction}
A codebook of supervised SRPs is generated from images of well known object categories expected to be present in a particular indoor scene (e.g., a microwave in a kitchen, a chair in a classroom). The codebook contains human-understandable elements which carry well-defined semantic meanings (similar to attributes \cite{farhadi2009describing} or object banks \cite{li2010object}).
In this regard, we introduce the first large-scale database of objects categories in indoor scenes (Sec. \ref{subsubsec:OCISdatabase}). The introduced database includes an extensive set of indoor objects (more than $1300$). The codebook of supervised SRPs is generated from images of the database by extracting dense mid-level patches after re-sizing the smallest dimension of each image to $256$ pixels. The number of SRPs in the compiled codebook is equal to the object categories in the OCIS database. For this purpose, in the feature space, each SRP is a max-pooled version of convolutional activations (Sec~\ref{subsec:feat_extraction}) of all the mid-level patches extracted from that object category.
The supervised codebook is then used in Sec. \ref{subsec:encoding_feat} to characterize a given scene image in terms of its constituent objects.

%--------------------------------------------------------------------
\subsubsection{Unsupervised SRPs}
\label{Unsupervised SRPs Extraction}
The codebook of unsupervised SRPs is generated from the patches extracted from the training data.
First, we densely and uniformly extract patches from training images by following the procedure described in Sec. \ref{Dense Patch Extraction}.
The SRPs can then be generated from these patches using any unsupervised clustering technique.
However, in our case, we randomly sample the patches as our unsupervised SRPs.
This is because, we are dealing with a very large number of extracted patches and an unsupervised clustering can be computationally prohibitive.
We demonstrate in our experiments (Sec.~\ref{subsec:results}, \ref{Ablative Analysis}) that random sampling does not cause any noticeable performance degradation, while achieving significant computational advantages.

Ideally, the codebook of SRPs should be all-inclusive and cover all discriminative aspects of indoor scenes. One might therefore expect a large number of SRPs in order to cover all the possible aspects of various scene categories.
While this is indeed the case, feature encoding from a single large codebook would be computationally burdensome (Sec. \ref{Ablative Analysis}).
We therefore propose to generate multiple codebooks of relatively smaller sizes. The association vectors from each of these codebooks can then be concatenated to generate a high dimensional feature vector. This guarantees the incorporation of a large number of SRPs at a low computational cost. To this end, we generate three unsupervised codebooks, each with $3000$ SRPs.
The codebook size was selected empirically on a small validation set.

The SRPs in the supervised codebook are semantically meaningful, however, they do not include all possible aspects of the different scene categories. The unsupervised codebook compensates this shortcoming and complements the supervised codebook. The combinations of both supervised and unsupervised codebooks, results therefore in an improved discrimination and accuracy (see Sec.~\ref{Ablative Analysis}).

%--------------------------------------------------------------------

%--------------------------------------------------------------------

%-------------------------------------------------------------------------
\subsection{Feature Encoding from SRPs}% Representations}
\label{subsec:encoding_feat}
Given an RGB image $I \in \mathbb{R} ^{H \times W \times 3}$, our task is to find its feature representation in terms of the previously generated codebooks of SRPs (Sec.~\ref{Supervised SRPs Extraction} and~\ref{Unsupervised SRPs Extraction}).
For this purpose, we first densely extract patches $\{x^{(i)}\in \mathbb{R}^{224 \times 224 \times 3}\}_{i=1}^{N}$ from the image using the procedure explained in Sec.~\ref{Dense Patch Extraction}.
Next, the patches are represented by their convolutional activations as discussed in Sec. \ref{subsec:feat_extraction}.
The patches are then encoded in terms of their association with the SRPs of the codebooks.
The following two strategies are devised for this purpose.

\subsubsection{Sparse Linear Coding}

Let $X \in \mathbb{R}^{4096 \times m}$ be a codebook of $m$ SRPs, a mid-level patch $x^{(i)}$ is sparsely reconstructed from the SRPs of the codebook using:
\begin{equation}
\min_{f^{(i)}} \left \| Xf^{(i)}-x^{(i)} \right \|_2 +\lambda \left \| f^{(i)} \right \|_1 .
\end{equation}
$\lambda$ is the regularization constant. The sparse coefficient vector $f^{(i)}$ is then used as the final feature representation of the patch.

\subsubsection{Proposed Classifier Similarity Metric Coding}

We propose a new soft encoding method which uses the maximum margin hyper-planes to measure feature associations. Given a codebook of $m$ SRPs, we train $m$ linear binary one-vs-all SVMs. An SVM finds the maximum margin hyperplane which optimally discriminates an SRP from all others. Let $W \in \mathbb{R} ^{4096 \times m}$ be the learnt weight matrix of all learnt SVMs. A patch $x^{(i)}$ can then be encoded in terms of the trained SVMs using: $f^{(i)}=W^Tx^{(i)}$.
Since we have multiple codebooks ($K$ in total), the patch $x^{(i)}$ is separately encoded from all of them. The final representation of $x^{(i)}$ is then achieved by concatenating the encoded feature representation from all codebooks into a single feature vector $f^{(i)}= \left [ f^{(i)}_1 f^{(i)}_2  \cdots f^{(i)}_K  \right ]$.

\subsection{Classification }

The encoded feature representations from all mid-level patches of the image are finally pooled to produce the overall feature representation of the image. Two commonly used pooling strategies (mean pooling and max pooling) are explored in our experiments (see Sec.~\ref{Ablative Analysis}). Finally, in order to perform classification, we use one-vs-one linear SVMs.
\begin{equation}
\underset{\mathbf{w}} {\operatorname{min}} \frac{1}{2}  \mathbf{w}\mathbf{w}^{T} + C \sum\limits_{i} \left( \text{max}(0, 1 - y^{(t)}\mathbf{w}^{T}{f}^{(i)})\right)^2.
\end{equation}

Where $\mathbf{w}$ is the normal vector to the learned max-margin hyper-plane, $C$ is the regularization parameter and $y^{(t)}$ is the binary class label of the feature vector ${f}^{(i)}$.

%-------------------------------------------------------------------------
\section{Experiments and Evaluation}

%-------------------------------------------------------------------------

We evaluate our approach on three indoor scene classification datasets. These include MIT-67 dataset, 15 Category Scene data set and NYU indoor scene dataset. Confusing inter-class similarities and high within-class variabilities make these datasets very challenging. Specifically, MIT-67 is the largest dataset of indoor scene images containing 67 classes. The images of many of these classes are very similar looking e.g., \emph{inside-subway} and \emph{inside-bus} (see Fig~\ref{fig:incorrect_Pred} for example confusing and challenging images).
Moreover, we also report results on two event and object classification datasets (Graz-02 dataset and 8-Sports event dataset) to demonstrate that the proposed technique is applicable to other related tasks.
 A detailed description of each of these datasets, followed by our experimental setups and the corresponding results are presented in Sec.~\ref{Scene Databases} and~\ref{subsec:results}.
 First, we provide a description of our introduced OCIS dataset below.

%---------------------------------------------------------------------------

\begin{SCfigure}[][t]
\centering
\includegraphics[width = 0.58\columnwidth]{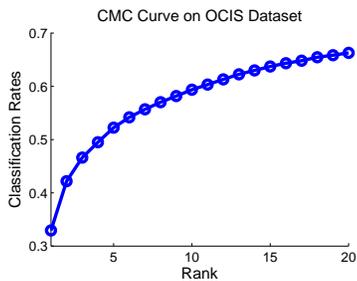}
\caption{CMC Curve for the benchmark evaluation on OCIS dataset. The curve illustrates the challenging nature of the dataset.}%A classification rate of only 32\% is achieved for rank-1 and 67\% for rank-20.}
\label{fig:CMC}
%\vspace{-1em}
\end{SCfigure}

\begin{figure}[b]
\centering
%\vspace{-1em}
\includegraphics[trim= 0 0 0 0em, clip, width=\columnwidth]{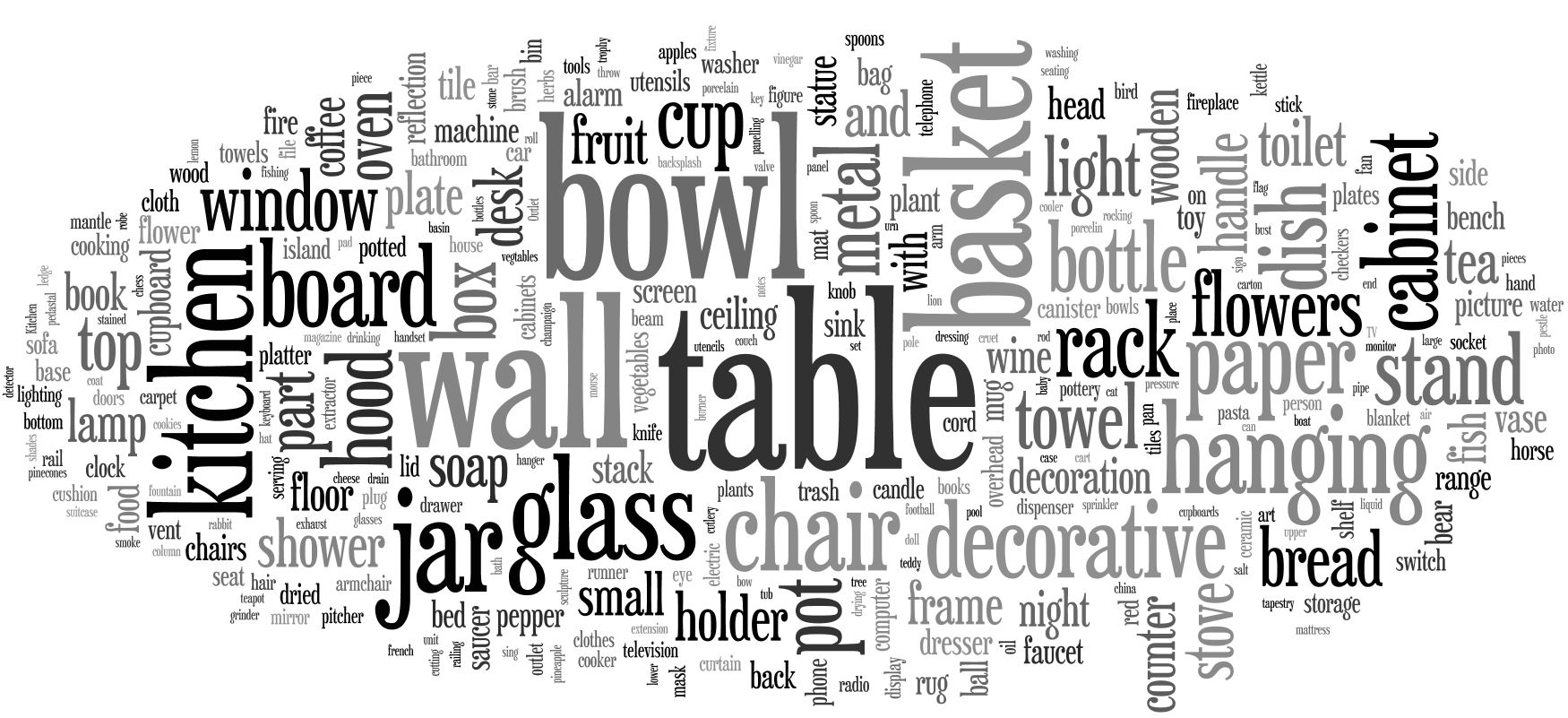}
\caption{A word cloud of the top 300 most frequently occurring classes in our introduced Object Categories in Indoor Scenes (OCIS) database.}
\label{fig:ClassNames}
%\vspace{-1em}
\end{figure}

%---------------------------------------------------------------
\begin{figure*}[t]
\centering
\includegraphics[width=\textwidth]{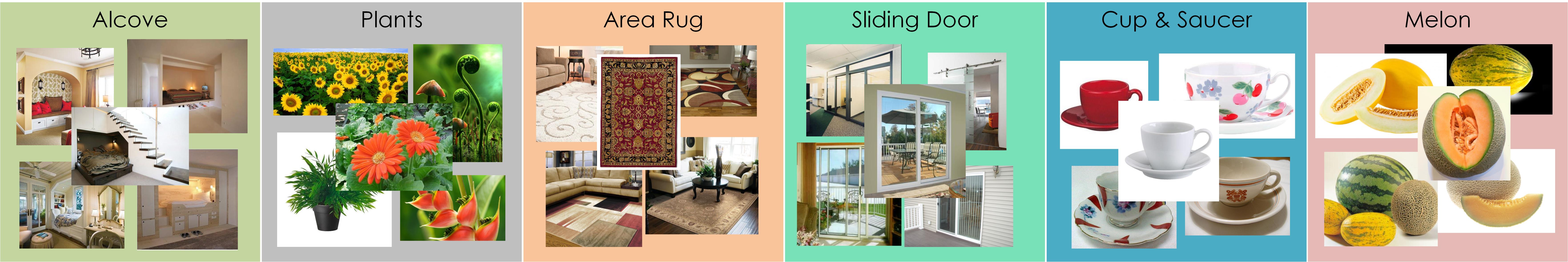}
%\vspace{-1.5em}
\caption{Example images from the `Object Categories in Indoor Scenes' dataset. This dataset contains a diverse set of object classes with different sizes and scales (e.g, \emph{Alcove} and \emph{Melon}). Each category includes a rich set of images with differences in appearance, shape, viewpoint and background.} %context.
%\vspace{-1em}
\label{fig:Dataset_Examples}
\end{figure*}
%---------------------------------------------------------------

%----------------------------------------------------------------
\begin{table*}
\centering
\scalebox{1.15}{
\begin{tabular}{@{}l c lc@{}}
\toprule
 \multicolumn{4}{c}{\textbf{MIT-67 Indoor Scenes Dataset}} \\
 \midrule
Method & Accuracy (\%) & Method & Accuracy (\%) \\
\midrule
ROI + GIST [CVPR'09] \cite{quattoni2009recognizing} & $26.1$ & OTC [ECCV'14] \cite{margolin2014otc} & $47.3$ \\

MM-Scene [NIPS'10] \cite{zhu2010large} & $28.3$  & Discriminative Patches [ECCV'12] \cite{singh2012unsupervised}
& $49.4$ \\

SPM [CVPR'06] \cite{lazebnik2006beyond} & $34.4$ & ISPR  [CVPR'14]\cite{linlearning14} & $50.1$ \\

Object Bank [NIPS'10] \cite{li2010object} & $37.6$& D-Parts [ICCV'13] \cite{sun2013learning} & $51.4$ \\

RBoW [CVPR'12] \cite{parizi2012reconfigurable} & $37.9$ & VC + VQ [CVPR'13] \cite{li2013harvesting} & $52.3$ \\

Weakly Supervised DPM [ICCV'11] \cite{pandey2011scene} & $43.1$  & IFV [CVPR'13]\cite{juneja2013blocks}  & $60.8$ \\

SPMSM [ECCV'12] \cite{kwitt2012scene} & $44.0$ & MLRep [NIPS'13] \cite{doersch2013mid} & $64.0$ \\

LPR-LIN [ECCV'12] \cite{sadeghi2012latent} & $44.8$ & CNN-MOP [ECCV'14]\cite{GongMOP14}  & $68.9$ \\

BoP [CVPR'13] \cite{juneja2013blocks} & $46.1$ & CNNaug-SVM [CVPRw'14] \cite{razavian2014cnn} & $69.0$ \\
\cmidrule{3-4}
Hybrid Parts + GIST + SP [ECCV'12] \cite{zheng2012learning} & $47.2$ & \textbf{Proposed DUCA} & \textbf{71.8} \\

\bottomrule
\end{tabular}
}
\caption{Mean accuracy on the MIT-67 Indoor Scenes Dataset. Comparisons with the previous state-of-the-art methods are also shown. Our approach performs best in comparison to techniques which use a single or multiple feature representations. }
\label{tab:MIT67acc}
%\vspace{-1em}
\end{table*}%----------------------------------------------------------------

\subsection{A Dataset of Object Categories in Indoor Scenes}
\label{subsubsec:OCISdatabase}
There is an exhaustive list of scene elements (including objects, structures and materials) that can be present in indoor scenes.
Any information about these scene elements can prove crucial for the scene categorization task (and even beyond - e.g., for the semantic labeling or attribute identification). However, to the best of our knowledge, there is no publicly available dataset of these indoor scene elements.
In this paper, we introduce the first large-scale OCIS (\emph{Object Categories in Indoor Scenes}) database.
The database contains a total of $15324$ images spanning more than $1300$ frequently occurring indoor object categories.
The number of images in each category is about $11$. The database can potentially be used for fine-grained scene categorization, high-level scene understanding and attribute based reasoning. In order to collect the data, a comprehensive list of $1325$ indoor objects was manually chosen from the labelings provided with the MIT-67 \cite{quattoni2009recognizing} dataset. This taxonomy includes a diverse set of objects classes ranging from a \emph{`house'} to a \emph{`handkerchief'}. A word cloud of the top 300 most frequently occurring classes is shown in Fig. \ref{fig:ClassNames}. The images for each class are then collected using an online image search (Google API). Each image contains one or more instances of a specific object category. In order to illustrate the diverse intra-class variability of this database, we show some example images in Fig.~\ref{fig:Dataset_Examples}. Our in-house annotated database will be made freely available to the research community.

For the benchmark evaluation, we represent the images of the database by their convolutional features and feed them to a linear classifier (SVM).
A train-test split of $66\%$-$33\%$ is defined for each class.
The classification results in terms of the Cumulative Match Curve (CMC) are shown in Fig.~\ref{fig:CMC}.
The rank-1 and rank-20 identification rates turn out to be only $32\%$ and $67\%$ respectively.
 These modest classification rates suggest that indoor object categorization is a very challenging task. % on the authors website.

%---------------------------------------------------------------------------
%Roberto: You use your own collected data (OCIS) to define the supervised codebooks, but then you use different databases for the actual classification (presumably making use of the OCIS supervised codebooks together with the unsupervised from the data, e.g. MIT-67). Are your features immune from database mismatch problems (between OCIS and MIT-67, etc.)?
%I am not sure if you need to say anything about this, all depends on whether this is common practice in image / vision processing. Certainly mismatch is a big deal in speech/speaker recognition.

\subsection{Evaluated Datasets}
\label{Scene Databases}

The performance of our proposed method is evaluated on MIT-67 dataset, 15 Category Scene data set, NYU Indoor Scene dataset, Graz-02 dataset and 8-Sports event dataset. Below, we present a brief description of each of these datasets followed by an analysis on the achieved performance.

\subsubsection{MIT-67 Dataset}
 It contains 15620 images of 67 indoor categories. For performance evaluation and comparison, we followed the standard evaluation protocol in \cite{quattoni2009recognizing} in which a subset of data is used (100 images per class) and a train-test split is defined to be $80\%-20\%$ for each class.

\subsubsection{15 Category Scene Dataset}
It contains images of 15 urban and natural scene categories. The number of images in each category ranges from 200-400. For our experiments, we use the same evaluation setup as in \cite{lazebnik2006beyond}, where 100 images per class are used for training and the rest for testing.

\subsubsection{NYU v1 Indoor Scene Dataset}
It consists of 7 indoor scene categories with a total of 2347 images. Following the standard experimental protocol \cite{silberman11indoor}, we used a $60\%-40\%$ train/test split for evaluation. Care has been taken while splitting the data to ensure that a minimal or no overlap of the consecutive frames exists between the training and testing sets.

\subsubsection{Inria Graz-02 Dataset}
It consists of 1096 images belonging to 3 classes (bikes, cars and people) in the presence of heavy clutter, occlusions and pose variations.
%The images of this dataset exhibit high intra-class variations.
For performance evaluation, we used the protocol defined in \cite{marszatek2007accurate}. Specifically, for each class, the first 150 odd images are used for training and the 150 even images are used for testing.

\subsubsection{UIUC 8-Sports Event Dataset}
It contains 1574 images of 8 sports categories. Following the protocol defined in \cite{li2007and}, we used 70 randomly sampled images for training and 60 for testing.

%----------------------------------------------------------------
\begin{table*}
\centering
\scalebox{1.15}{
\begin{tabular}{@{}lc lc@{}}
\toprule
 \multicolumn{4}{c}{\textbf{15 Category Scene Dataset}} \\
 \midrule
Method & Accuracy(\%) & Method & Accuracy (\%) \\
\midrule
 GIST-color [IJCV'01] \cite{oliva2001modeling} & $69.5$ &  ISPR [CVPR'14] \cite{linlearning14} & $85.1$ \\

RBoW [CVPR'12] \cite{parizi2012reconfigurable} & $78.6$ & VC + VQ [CVPR'13] \cite{li2013harvesting} & $85.4$ \\

Classemes [ECCV'10] \cite{torresani2010efficient} & $80.6$ & LMLF [CVPR'10] \cite{boureau2010learning} & $85.6$ \\

Object Bank [NIPS'10] \cite{li2010object} & $80.9$
& LPR-RBF [ECCV'12] \cite{sadeghi2012latent} & $85.8$ \\

SPM [CVPR'06] \cite{lazebnik2006beyond} & $81.4$ & Hybrid Parts + GIST + SP [ECCV'12] \cite{zheng2012learning} & $86.3$ \\

SPMSM [ECCV'12] \cite{kwitt2012scene} & $82.3$ &  CENTRIST+LCC+Boosting [CVPR'11] \cite{yuan2011mining}  & $87.8$ \\

LCSR [CVPR'12] \cite{shabou2012locality} &  $82.7$ & RSP [ECCV'12] \cite{jiang2012randomized} & $88.1$ \\

SP-pLSA [PAMI'08] \cite{bosch2008scene} & $83.7$ & IFV \cite{vedaldi08vlfeat} & $89.2$ \\

CENTRIST [PAMI'11] \cite{wu2011centrist} &  $83.9$ & LScSPM [CVPR'10] \cite{gao2010local} & $89.7$ \\

 HIK [ICCV'09]\cite{wu2009beyond} & $84.1$
& \\

\cmidrule{3-4}
OTC [ECCV'14] \cite{margolin2014otc} & $84.4$
& \textbf{Proposed DUCA} & \textbf{94.5} \\
\bottomrule
\end{tabular}
}
%\vspace{-0.5em}
\caption{Mean accuracy on the 15 Category Scene Dataset. Comparisons with the previous best techniques are also shown.}
\label{tab:15scenes_acc}
%\vspace{-1em}
\end{table*}%----------------------------------------------------------------

%----------------------------------------------------------------
\begin{table}
\centering
\scalebox{1.15}{
\begin{tabular}{@{}lc@{}}
\toprule
 \multicolumn{2}{c}{\textbf{UIUC 8-Sports Dataset}} \\
 \midrule
Method & Accuracy (\%) \\
\midrule
GIST-color [IJCV'01] \cite{oliva2001modeling} & $70.7$ \\

MM-Scene [NIPS'10] \cite{zhu2010large} & $71.7$ \\

Graphical Model [ICCV'07] \cite{li2007and} & $73.4$ \\

Object Bank [NIPS'10] \cite{li2010object} & $76.3$ \\

Object Attributes [ECCV'12] \cite{li2012objects} & $77.9$ \\

CENTRIST [PAMI'11] \cite{wu2011centrist} &  $78.2$ \\

RSP [ECCV'12] \cite{jiang2012randomized} & $79.6$ \\

SPM [CVPR'06] \cite{lazebnik2006beyond} & $81.8$ \\

SPMSM [ECCV'12] \cite{kwitt2012scene} & $83.0$ \\

Classemes [ECCV'10] \cite{torresani2010efficient} & $84.2$ \\

HIK [ICCV'09] \cite{wu2009beyond} & $84.2$ \\

LScSPM [CVPR'10] \cite{gao2010local} & $85.3$ \\

LPR-RBF [ECCV'12] \cite{sadeghi2012latent} & $86.2$ \\

Hybrid Parts + GIST + SP [ECCV'12] \cite{zheng2012learning} & $87.2$ \\

LCSR [CVPR'12] \cite{shabou2012locality} &  $87.2$ \\

VC + VQ [CVPR'13] \cite{li2013harvesting} & $88.4$ \\

IFV \cite{vedaldi08vlfeat} & $90.8$ \\

ISPR [CVPR'14] \cite{linlearning14} & $89.5$ \\
\midrule
\textbf{Proposed DUCA} & \textbf{98.7} \\
\bottomrule
\end{tabular}
}
%\vspace{-0.5em}
\caption{Mean accuracy on the UIUC 8-Sports Dataset. }
\label{tab:Sports_acc}
%\vspace{-1em}
\end{table}%----------------------------------------------------------------

% Most results taken from ISPR paper

%-------------------------------------------------------------------------
\subsection{Experimental Results}\label{subsec:results}
The quantitative results of the proposed method for the task of indoor scene categorization are presented in Tables \ref{tab:MIT67acc}, \ref{tab:15scenes_acc} and \ref{tab:nyu_acc}.
The proposed method achieves the highest classification rate on all three datasets. Compared with the existing state of the art, a relative performance increment of 4.1\%, 5.4\% and 1.3\% is achieved for MIT-67, Scene-15 and NYU datasets respectively. Amongst the compared methods, the mid-level feature representation based methods \cite{doersch2013mid, singh2012unsupervised, sun2013learning} perform better than the others. Our proposed mid-level features based method not only outperforms their accuracy but is also computationally efficient (e.g., \cite{doersch2013mid} takes weeks to train several part detectors). Furthermore, once compared with existing methods, our proposed method uses a lower dimensional feature representation for classification (e.g., the Juneja et al. \cite{juneja2013blocks} Improved Fisher Vector (IFV) has dimensionality $> 200$K; the Gong et al. \cite{GongMOP14} MOP representation has $> 12$K).

In addition to indoor scene classification, we also evaluate our approach on other scene classification tasks where large variations and deformations are present.
To this end, we report the classification results on the UIUC 8-Sports dataset and the Graz-02 dataset (see Tables \ref{tab:Sports_acc} and \ref{tab:graz_acc}).
It is interesting to note that the Graz-02 dataset contains heavy clutter, pose and scale variations (e.g., for some `car' images only 5\% of the pixels are covered by the \emph{car} in a scene).
Our approach achieved high accuracies of $98.7\%$ and $98.6\%$ respectively on the UIUC 8-Sports and Graz-02 datasets.
These performances are $10.2\%$ and $12.6\%$ higher than the previous best methods on UIUC 8-Sports and Graz-02 datasets respectively.

%----------------------------------------------------------------
\begin{table}
\centering
\scalebox{1.15}{
\begin{tabular}{@{}lc@{}}
\toprule
 \multicolumn{2}{c}{\textbf{NYU Indoor Scenes Dataset}} \\
 \midrule
Method & Accuracy (\%) \\
\midrule
BoW-SIFT [ICCVw'11] \cite{silberman11indoor} & $55.2$ \\
RGB-LLC [TC'13] \cite{tao2013rank} & $78.1$ \\
RGB-LLC-RPSL [TC'13] \cite{tao2013rank} & $79.5$ \\
\midrule
\textbf{Proposed DUCA} & $\mathbf{80.6}$\\
\bottomrule
\end{tabular}
}
%\vspace{-0.5em}
\caption{Mean Accuracy for the NYU v1 Dataset. }
\label{tab:nyu_acc}
%\vspace{-1em}
\end{table}
%----------------------------------------------------------------

%----------------------------------------------------------------
\begin{table}
\centering
\scalebox{1.1}{
\begin{tabular}{@{}l ccc @{\hspace{2em}}c@{}}
\toprule
 \multicolumn{5}{c}{\textbf{Graz-02 Dataset}} \\
 \midrule
& Cars & People & Bikes & Overall \\
\midrule
OLB [SCIA'05] \cite{opelt2005object} & 70.7 & 81.0
 & 76.5  & 76.1 \\
 VQ [ICCV'07] \cite{tuytelaars2007vector} & 80.2 & 85.2 & 89.5 & 85.0 \\
ERC-F [PAMI'08] \cite{moosmann2008randomized} & 79.9 & - & 84.4 & {82.1}\\
TSD-IB [BMVC'11] \cite{krapac2011learning} & 87.5 & 85.3 & 91.2 &  88.0 \\
TSD-k [BMVC'11] \cite{krapac2011learning} & 84.8 & 87.3 & 90.7 & 87.6 \\
\midrule
\textbf{Proposed DUCA} & 98.7 & 98.0 &   99.0 & \textbf{98.6} \\
\bottomrule
\end{tabular}
}
%\vspace{-0.5em}
\caption{Equal Error Rates (EER) for the Graz-02 dataset. }
\label{tab:graz_acc}
%\vspace{-1em}
\end{table}
%----------------------------------------------------------------

\begin{figure}
\centering
\includegraphics[trim=51em 28.2em 14em 27.5em , clip, width=1\columnwidth]{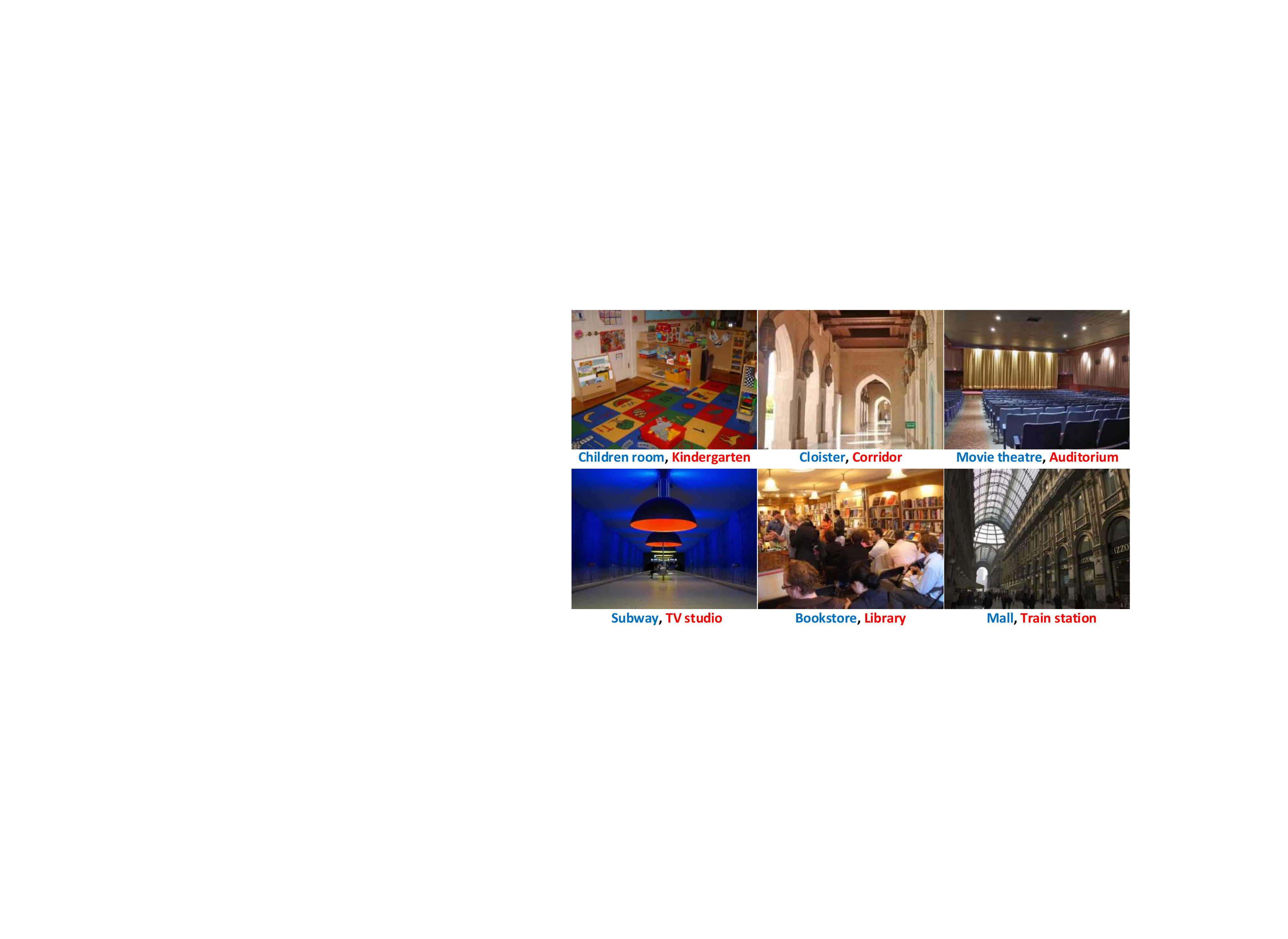}
%\vspace{-1.5em}
\caption{Examples mistakes and the limitations of our method. Most of the incorrect predictions are due to ambiguous cases. The actual and predicted class names are shown in \emph{`blue'} and \emph{`red'} respectively. \emph{(Best viewed in color)}}
%\vspace{-0.5em}
\label{fig:incorrect_Pred}
%\vspace{-0.5em}
\end{figure}

\begin{figure*}[htp!]
\centering
\begin{subfigure}[]{0.27\textwidth}
\imagebox{40mm}{\includegraphics[width=\textwidth]{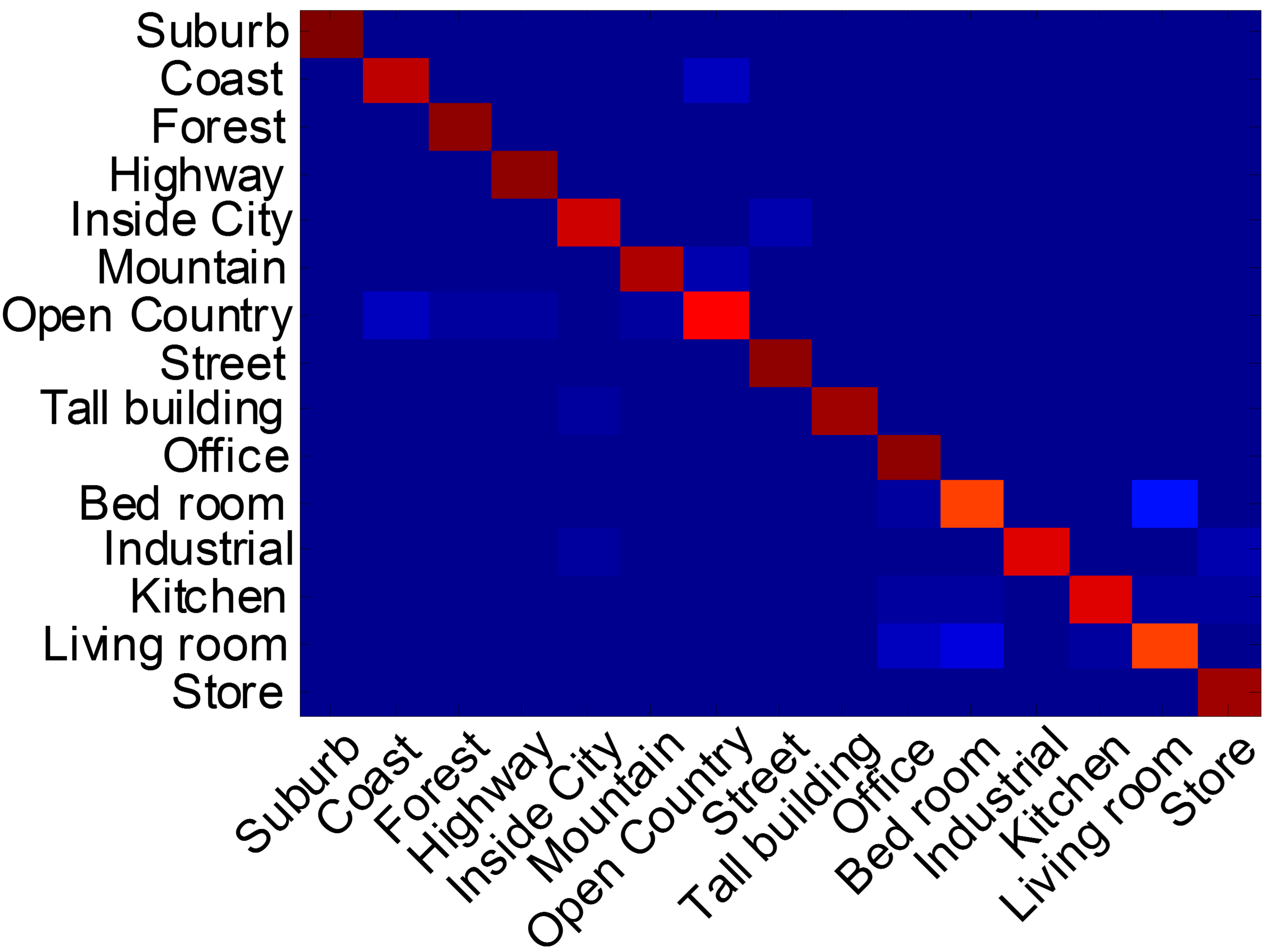}}
%\vspace{-1em}
\caption{15 Categories Scenes Dataset}
\label{1}
\end{subfigure}
\quad
\begin{subfigure}[]{0.27\textwidth}
\imagebox{39mm}{\includegraphics[width=\textwidth]{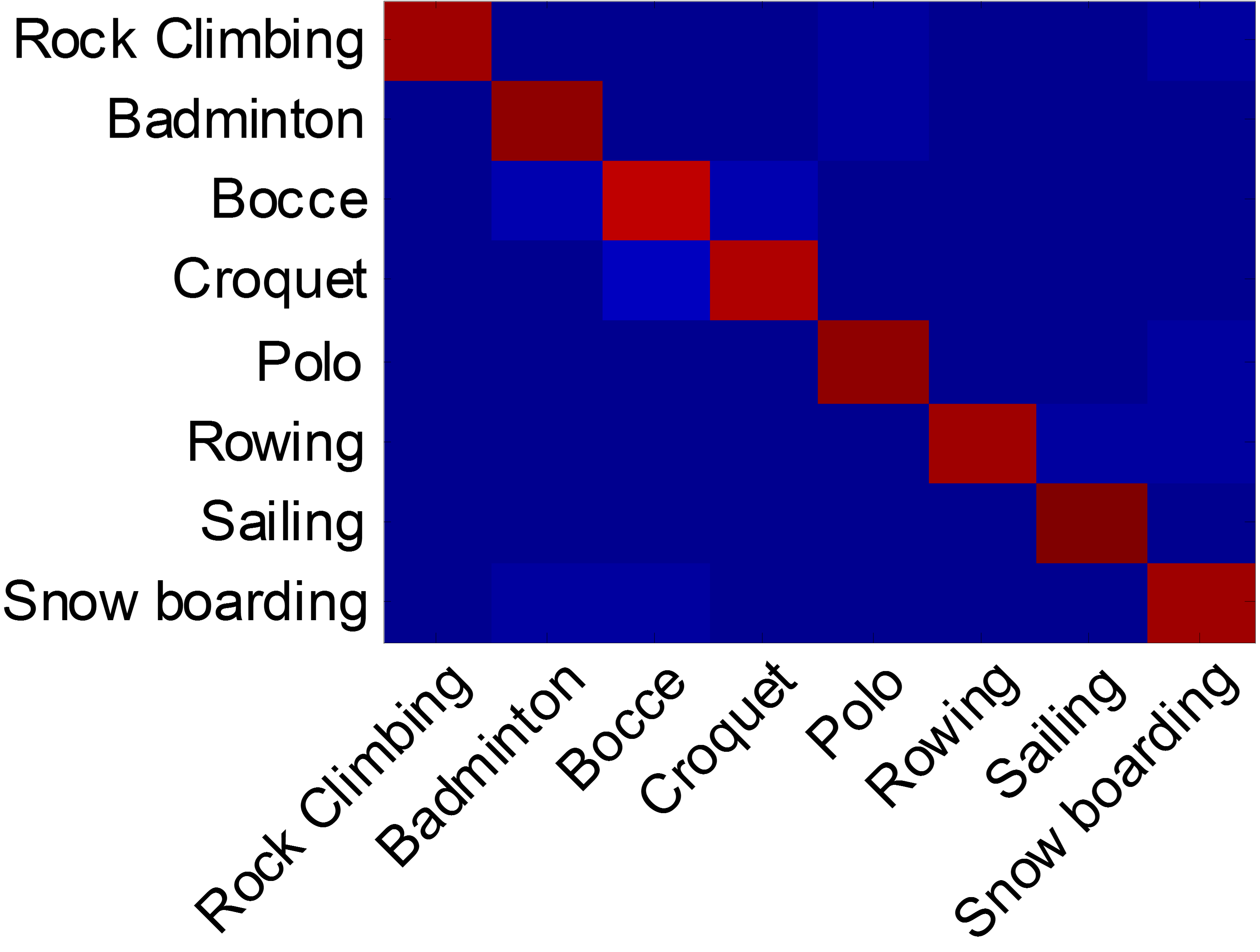}}
%\vspace{-1em}
\caption{UIUC 8-Sports Dataset}
\label{2}
\end{subfigure}
\quad
\begin{subfigure}[]{0.31\textwidth}
\imagebox{40mm}{\includegraphics[width=\textwidth]{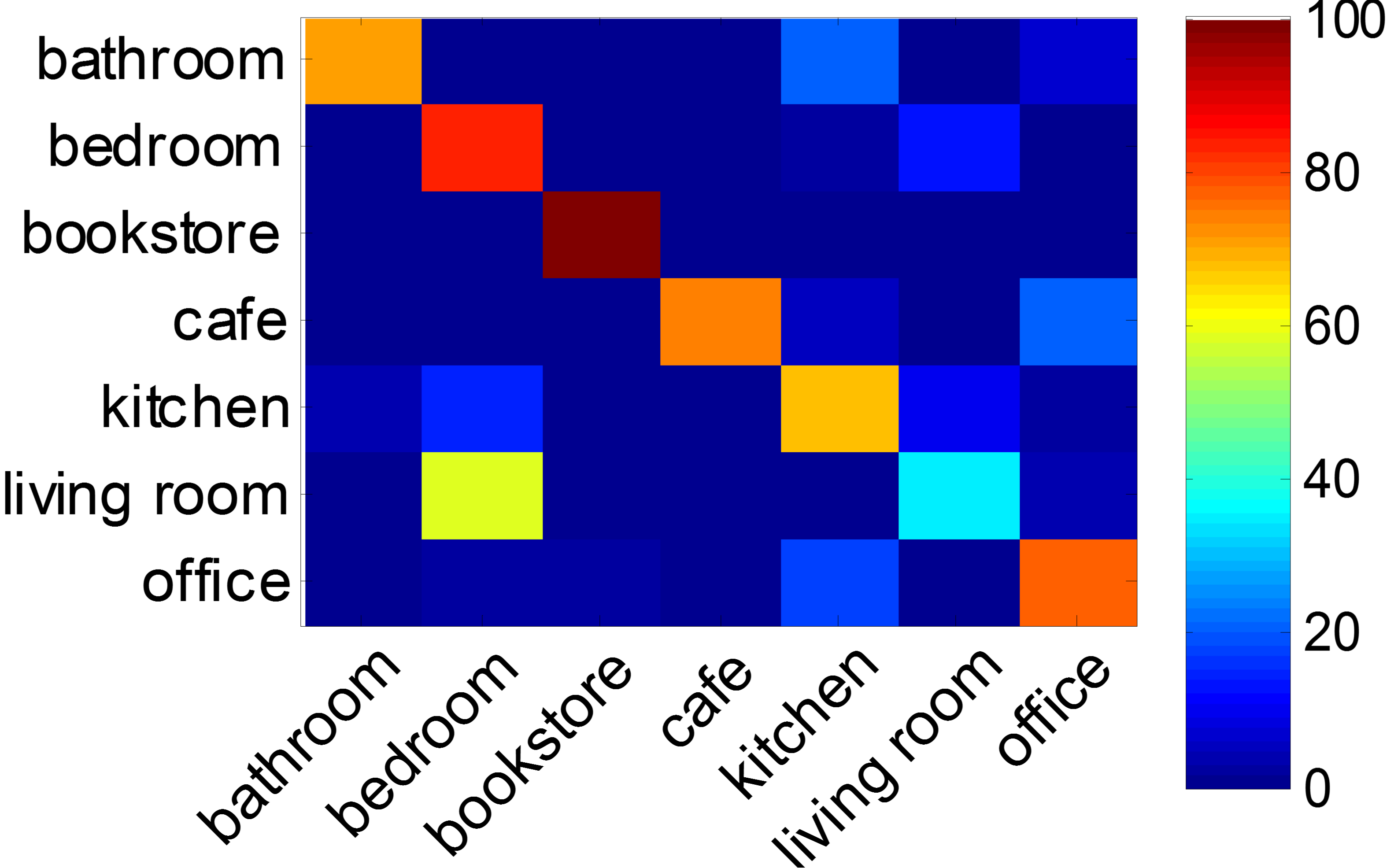}}
%\vspace{-1em}
\caption{NYU Indoor Scenes Dataset}
\label{3}
\end{subfigure}
%\vspace{-0.5em}
\caption{Confusion matrices for three scene classification datasets. \emph{(Best viewed in color)}}
\label{fig:confu_mat}
%\vspace{-0.5em}
\end{figure*}

\begin{figure*}
\centering
\includegraphics[trim = 0 38em 0 38em, clip, width=1\textwidth]{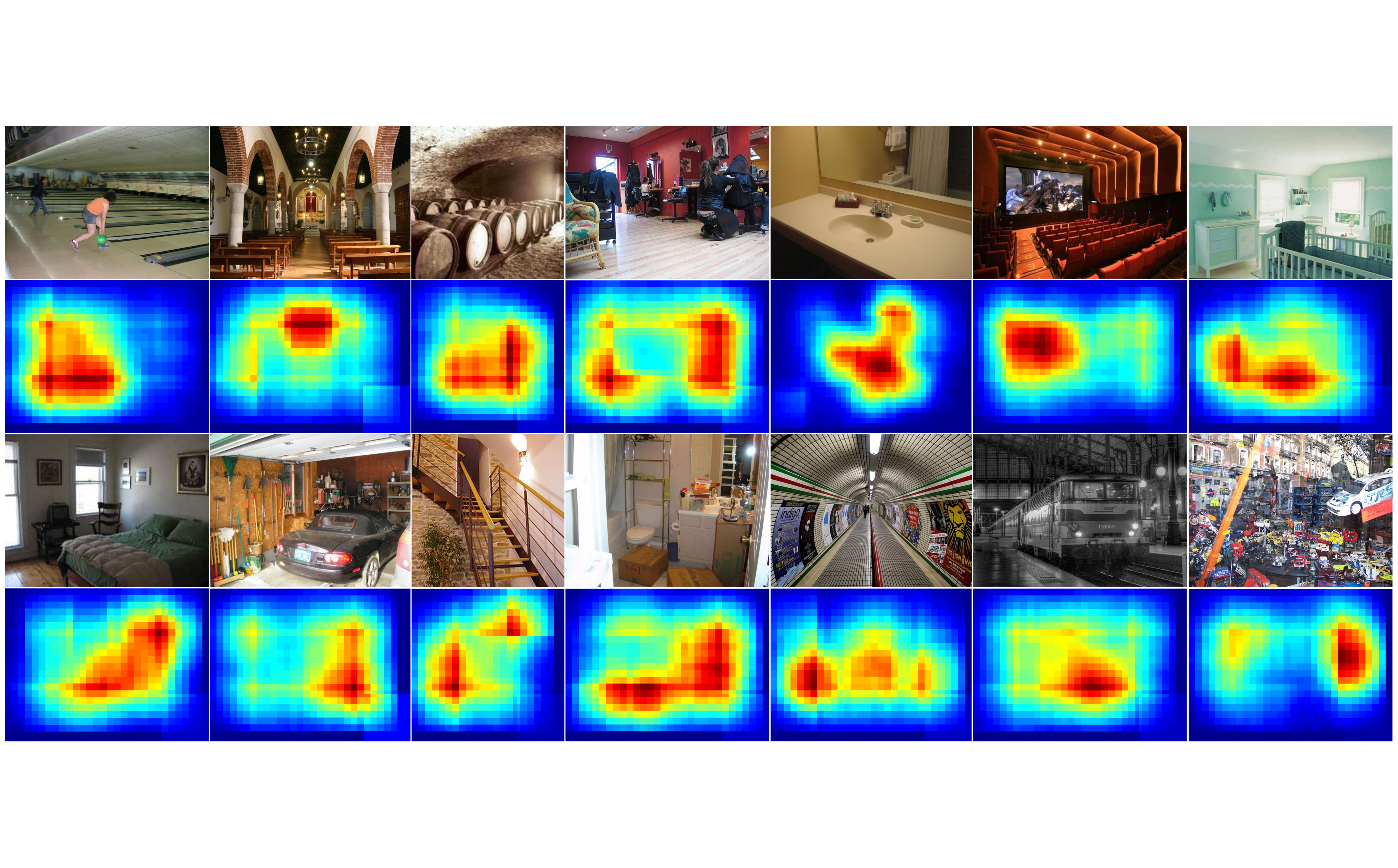}
%\vspace{-1.5em}
\caption{The contribution of distinctive patches for the correct class prediction of a scene are shown in the form of a heat map (`\emph{red}' means more contribution). These examples show that our approach captures the discriminative properties of distinctive mid-level patches and uses them to predict the correct class. \emph{(Best viewed in color)}}
\label{fig:discr_patches}
%\vspace{-1em}
\end{figure*}

\begin{figure*}[t]
\centering
\includegraphics[width=0.8\linewidth]{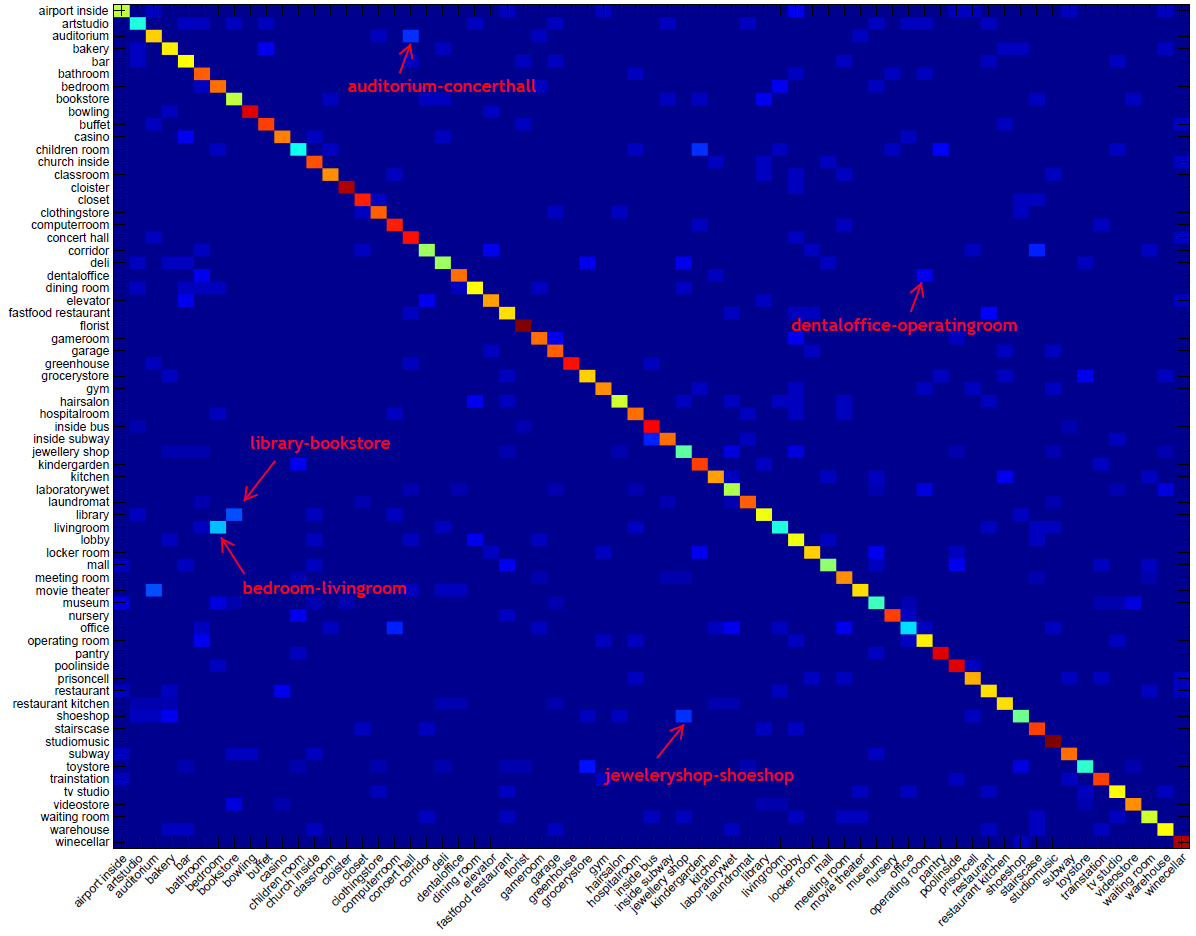}
% trim= 45mm 0mm  20mm  10mm with pdf
\caption{Confusion Matrix for MIT-67 dataset. It can be seen that the proposed method confuses similar looking classes with each others e.g., library with bookstore; bedroom with living room and dental office with operating room. \emph{(Best viewed in color)}}
\label{fig:MIT67_CM}
\end{figure*}

The class-wise classification accuracies of the MIT-67, UIUC 8-Sports, Scene-15 and NYU datasets are shown in the form of confusion matrices in Fig.~\ref{fig:confu_mat} and \ref{fig:MIT67_CM} respectively.
Note the very strong diagonal in all confusion matrices.
The majority ($>90\%$) of the mistakes are made for the closely related classes e.g., \emph{coast-opencountry} (Fig. \ref{1}), \emph{croquet-bocce} (Fig. \ref{2}), \emph{bedroom-livingroom} (Fig. \ref{3}), \emph{dentaloffice-operatingroom} (Fig. \ref{fig:MIT67_CM}) and \emph{library-bookstore} (Fig. \ref{fig:MIT67_CM}).
We also show examples of miss-classified images in Fig.~\ref{fig:incorrect_Pred}.
The results show that the classes with significant visual and semantic similarities are confused amongst each others e.g., \emph{childrenroom-kindergarten} and \emph{movietheatre-auditorium} (Fig.~\ref{fig:incorrect_Pred}).

In order to visualize which patches contributed most towards a correct classification, we plot the heat map of the patch contribution scores in Fig. \ref{fig:discr_patches}. It turns out that the most distinctive patches, which carry valuable information, have a higher contribution towards the correct prediction of a scene class.
{\color{black} Moreover, mid-level patches carry an intermediate level of scene details and contextual relationships between objects which help in the scene classification process. }

\subsection{Ablative Analysis}
\label{Ablative Analysis}
To analyze the effect of the different components of the proposed scheme on the final performance, we conduct an ablation study.
Table \ref{tab:ablation} summarizes the results achieved on the MIT-67 Scenes dataset when different components were replaced or removed from the final framework.

It turns out that the supervised and unsupervised codebooks individually perform reasonably well. However, their combination gives state of the art performance.
For the unsupervised codebook, k-means clustering performs slightly better however, at the cost of a considerable amount of computational resources ($\sim 40$GB RAM for the MIT-67 dataset) and processing time ($\sim1$  day for the MIT-67 dataset).
In contrast, the random sampling of MRPs gives a comparable performance with a big boost in computational efficiency.
%The feature encoding step becomes efficient with the use of multiple smaller codebooks.
Feature encoding from a single large codebook does not only produce a lower performance, but it also requires more computational time and memory. In our experiments, feature encoding from one single large codebook requires almost twice the time ($\sim45$ sec/image) taken by multiple smaller codebooks ($\sim25$ sec/image).
The resulting features performed best when the max-pooling operation was applied to combine them.

%--------------------------------------------------------------------
\begin{table}
\centering
\scalebox{1.15}{
\begin{tabular}{lc}
\toprule
Variants of Our Approach & Accuracy (\%) \\
\midrule
Supervised codebook  & $68.5$ \\
Unsupervised codebook & $69.9$ \\
Supervised + Unsupervised & $71.8$\\
\midrule
K-means clustering & $72.0$\\
Random sampling & $71.8$\\
\midrule
Single large codebook & $71.4$\\
Multiple smaller codebooks & $71.8$\\
\midrule
Sparse linear coding & $71.8$\\
Classifier similarity metric coding  & $69.9$\\
\midrule
Mean-poling & $69.7$\\
Max-pooling & $71.8$\\
\midrule
Original data & $69.1$ \\
Data augmentation & $71.8$\\
\bottomrule
\end{tabular}
}
%\vspace{-0.5em}
\caption{Ablative Analysis on MIT-67 Scene Dataset. }
\label{tab:ablation}
%\vspace{-2em}
\end{table}
%--------------------------------------------------------------------

%-------------------------------------------------------------------------
\section{Conclusion}

 This paper proposed a robust feature representation based on discriminative mid-level convolutional activations, for highly variable indoor scenes. To suitably contrive the convolutional activations for indoor scenes, the paper proposed to break their inherently preserved global spatial structure by encoding them in a number of of multiple codebooks.
These codebooks are composed of distinctive patches and of the semantically labeled elements.
For the labeled elements, we introduced the first large-scale dataset of object categories of indoor scenes.
Our approach achieves state-of-the-art performance on five very challenging datasets.

%----------------------------------------------------------------------
\section*{Acknowledgements}
This research was supported by the SIRF and IPRS scholarships from the University of Western Australia (UWA) and the Australian Research Council (ARC) grants DP110102166, DP150100294 and DE120102960.
%We gratefully acknowledge the support of NVIDIA Corporation with the donation of the Tesla K40 GPU used for this research.

\bibliographystyle{IEEEtrans}
\bibliography{egbib}

% Generated by IEEEtranS.bst, version: 1.13 (2008/09/30)
\begin{thebibliography}{10}
\providecommand{\url}[1]{#1}
\csname url@samestyle\endcsname
\providecommand{\newblock}{\relax}
\providecommand{\bibinfo}[2]{#2}
\providecommand{\BIBentrySTDinterwordspacing}{\spaceskip=0pt\relax}
\providecommand{\BIBentryALTinterwordstretchfactor}{4}
\providecommand{\BIBentryALTinterwordspacing}{\spaceskip=\fontdimen2\font plus
\BIBentryALTinterwordstretchfactor\fontdimen3\font minus
  \fontdimen4\font\relax}
\providecommand{\BIBforeignlanguage}[2]{{%
\expandafter\ifx\csname l@#1\endcsname\relax
\typeout{** WARNING: IEEEtranS.bst: No hyphenation pattern has been}%
\typeout{** loaded for the language `#1'. Using the pattern for}%
\typeout{** the default language instead.}%
\else
\language=\csname l@#1\endcsname
\fi
#2}}
\providecommand{\BIBdecl}{\relax}
\BIBdecl

\bibitem{bosch2008scene}
A.~Bosch, A.~Zisserman, and X.~Muoz, ``Scene classification using a hybrid
  generative/discriminative approach,'' \emph{Transactions on Pattern Analysis
  and Machine Intelligence}, vol.~30, no.~4, pp. 712--727, 2008.

\bibitem{boureau2010learning}
Y.-L. Boureau, F.~Bach, Y.~LeCun, and J.~Ponce, ``Learning mid-level features
  for recognition,'' in \emph{International Conference on Computer Vision and
  Pattern Recognition}.\hskip 1em plus 0.5em minus 0.4em\relax IEEE, 2010, pp.
  2559--2566.

\bibitem{Chatfield14}
K.~Chatfield, K.~Simonyan, A.~Vedaldi, and A.~Zisserman, ``Return of the devil
  in the details: Delving deep into convolutional nets,'' in \emph{British
  Machine Vision Conference}, 2014.

\bibitem{doersch2013mid}
C.~Doersch, A.~Gupta, and A.~A. Efros, ``Mid-level visual element discovery as
  discriminative mode seeking,'' in \emph{Advances in Neural Information
  Processing Systems}, 2013, pp. 494--502.

\bibitem{farhadi2009describing}
A.~Farhadi, I.~Endres, D.~Hoiem, and D.~Forsyth, ``Describing objects by their
  attributes,'' in \emph{International Conference on Computer Vision and
  Pattern Recognition}.\hskip 1em plus 0.5em minus 0.4em\relax IEEE, 2009, pp.
  1778--1785.

\bibitem{fei2005bayesian}
L.~Fei-Fei and P.~Perona, ``A bayesian hierarchical model for learning natural
  scene categories,'' in \emph{International Conference on Computer Vision and
  Pattern Recognition}, vol.~2.\hskip 1em plus 0.5em minus 0.4em\relax IEEE,
  2005, pp. 524--531.

\bibitem{gao2010local}
S.~Gao, I.~W. Tsang, L.-T. Chia, and P.~Zhao, ``Local features are not
  lonely--laplacian sparse coding for image classification,'' in
  \emph{International Conference on Computer Vision and Pattern
  Recognition}.\hskip 1em plus 0.5em minus 0.4em\relax IEEE, 2010, pp.
  3555--3561.

\bibitem{GongMOP14}
Y.~Gong, L.~Wang, R.~Guo, and S.~Lazebnik, ``Multi-scale orderless pooling of
  deep convolutional activation features,'' in \emph{European Conference on
  Computer Vision}.\hskip 1em plus 0.5em minus 0.4em\relax Springer
  International Publishing, 2014, pp. 392--407.

\bibitem{gupta2013perceptual}
S.~Gupta, P.~Arbelaez, and J.~Malik, ``Perceptual organization and recognition
  of indoor scenes from rgb-d images,'' in \emph{International Conference on
  Computer Vision and Pattern Recognition}.\hskip 1em plus 0.5em minus
  0.4em\relax IEEE, 2013, pp. 564--571.

\bibitem{he2014spatial}
K.~He, X.~Zhang, S.~Ren, and J.~Sun, ``Spatial pyramid pooling in deep
  convolutional networks for visual recognition,'' in \emph{European Conference
  on Computer Vision}.\hskip 1em plus 0.5em minus 0.4em\relax Springer, 2014,
  pp. 346--361.

\bibitem{jia2014caffe}
Y.~Jia, E.~Shelhamer, J.~Donahue, S.~Karayev, J.~Long, R.~Girshick,
  S.~Guadarrama, and T.~Darrell, ``Caffe: Convolutional architecture for fast
  feature embedding,'' \emph{arXiv preprint arXiv:1408.5093}, 2014.

\bibitem{jiang2012randomized}
Y.~Jiang, J.~Yuan, and G.~Yu, ``Randomized spatial partition for scene
  recognition,'' in \emph{European Conference on Computer Vision}.\hskip 1em
  plus 0.5em minus 0.4em\relax Springer, 2012, pp. 730--743.

\bibitem{juneja2013blocks}
M.~Juneja, A.~Vedaldi, C.~Jawahar, and A.~Zisserman, ``Blocks that shout:
  Distinctive parts for scene classification,'' in \emph{International
  Conference on Computer Vision and Pattern Recognition}.\hskip 1em plus 0.5em
  minus 0.4em\relax IEEE, 2013, pp. 923--930.

\bibitem{krapac2011learning}
J.~Krapac, J.~Verbeek, F.~Jurie \emph{et~al.}, ``Learning tree-structured
  descriptor quantizers for image categorization,'' in \emph{British Machine
  Vision Conference}, 2011.

\bibitem{krizhevsky2012imagenet}
A.~Krizhevsky, I.~Sutskever, and G.~E. Hinton, ``Imagenet classification with
  deep convolutional neural networks,'' in \emph{Advances in Neural Information
  Processing Systems}, 2012, pp. 1097--1105.

\bibitem{kumar2005hierarchical}
S.~Kumar and M.~Hebert, ``A hierarchical field framework for unified
  context-based classification,'' in \emph{Computer Vision, 2005. ICCV 2005.
  Tenth IEEE International Conference on}, vol.~2.\hskip 1em plus 0.5em minus
  0.4em\relax IEEE, 2005, pp. 1284--1291.

\bibitem{kwitt2012scene}
R.~Kwitt, N.~Vasconcelos, and N.~Rasiwasia, ``Scene recognition on the semantic
  manifold,'' in \emph{European Conference on Computer Vision}.\hskip 1em plus
  0.5em minus 0.4em\relax Springer, 2012, pp. 359--372.

\bibitem{lazebnik2006beyond}
S.~Lazebnik, C.~Schmid, and J.~Ponce, ``Beyond bags of features: Spatial
  pyramid matching for recognizing natural scene categories,'' in
  \emph{International Conference on Computer Vision and Pattern Recognition},
  vol.~2.\hskip 1em plus 0.5em minus 0.4em\relax IEEE, 2006, pp. 2169--2178.

\bibitem{li2007and}
L.-J. Li and L.~Fei-Fei, ``What, where and who? classifying events by scene and
  object recognition,'' in \emph{International Conference on Computer
  Vision}.\hskip 1em plus 0.5em minus 0.4em\relax IEEE, 2007, pp. 1--8.

\bibitem{li2010object}
L.-J. Li, H.~Su, L.~Fei-Fei, and E.~P. Xing, ``Object bank: A high-level image
  representation for scene classification \& semantic feature sparsification,''
  in \emph{Advances in Neural Information Processing Systems}, 2010, pp.
  1378--1386.

\bibitem{li2012objects}
L.-J. Li, H.~Su, Y.~Lim, and L.~Fei-Fei, ``Objects as attributes for scene
  classification,'' in \emph{Trends and Topics in Computer Vision}.\hskip 1em
  plus 0.5em minus 0.4em\relax Springer, 2012, pp. 57--69.

\bibitem{li2013harvesting}
Q.~Li, J.~Wu, and Z.~Tu, ``Harvesting mid-level visual concepts from
  large-scale internet images,'' in \emph{International Conference on Computer
  Vision and Pattern Recognition}.\hskip 1em plus 0.5em minus 0.4em\relax IEEE,
  2013, pp. 851--858.

\bibitem{linlearning14}
D.~Lin, C.~Lu, R.~Liao, and J.~Jia, ``Learning important spatial pooling
  regions for scene classification,'' 2014.

\bibitem{margolin2014otc}
R.~Margolin, L.~Zelnik-Manor, and A.~Tal, ``Otc: A novel local descriptor for
  scene classification,'' in \emph{European Conference on Computer
  Vision}.\hskip 1em plus 0.5em minus 0.4em\relax Springer, 2014, pp. 377--391.

\bibitem{marszatek2007accurate}
M.~Marszatek and C.~Schmid, ``Accurate object localization with shape masks,''
  in \emph{International Conference on Computer Vision and Pattern
  Recognition}.\hskip 1em plus 0.5em minus 0.4em\relax IEEE, 2007, pp. 1--8.

\bibitem{moosmann2008randomized}
F.~Moosmann, E.~Nowak, and F.~Jurie, ``Randomized clustering forests for image
  classification,'' \emph{Transactions on Pattern Analysis and Machine
  Intelligence}, vol.~30, no.~9, pp. 1632--1646, 2008.

\bibitem{nowak2006sampling}
E.~Nowak, F.~Jurie, and B.~Triggs, ``Sampling strategies for bag-of-features
  image classification,'' in \emph{European Conference on Computer
  Vision}.\hskip 1em plus 0.5em minus 0.4em\relax Springer, 2006, pp. 490--503.

\bibitem{oliva2001modeling}
A.~Oliva and A.~Torralba, ``Modeling the shape of the scene: A holistic
  representation of the spatial envelope,'' \emph{International Journal of
  Computer Vision}, vol.~42, no.~3, pp. 145--175, 2001.

\bibitem{opelt2005object}
A.~Opelt and A.~Pinz, ``Object localization with boosting and weak supervision
  for generic object recognition,'' in \emph{SCIA}.\hskip 1em plus 0.5em minus
  0.4em\relax Springer, 2005, pp. 862--871.

\bibitem{oquab2013learning}
M.~Oquab, L.~Bottou, I.~Laptev, J.~Sivic \emph{et~al.}, ``Learning and
  transferring mid-level image representations using convolutional neural
  networks,'' 2014.

\bibitem{ouyang2014deepid}
W.~Ouyang, P.~Luo, X.~Zeng, S.~Qiu, Y.~Tian, H.~Li, S.~Yang, Z.~Wang, Y.~Xiong,
  C.~Qian \emph{et~al.}, ``Deepid-net: multi-stage and deformable deep
  convolutional neural networks for object detection,'' \emph{arXiv preprint
  arXiv:1409.3505}, 2014.

\bibitem{pandey2011scene}
M.~Pandey and S.~Lazebnik, ``Scene recognition and weakly supervised object
  localization with deformable part-based models,'' in \emph{International
  Conference on Computer Vision}.\hskip 1em plus 0.5em minus 0.4em\relax IEEE,
  2011, pp. 1307--1314.

\bibitem{parizi2012reconfigurable}
S.~N. Parizi, J.~G. Oberlin, and P.~F. Felzenszwalb, ``Reconfigurable models
  for scene recognition,'' in \emph{International Conference on Computer Vision
  and Pattern Recognition}.\hskip 1em plus 0.5em minus 0.4em\relax IEEE, 2012,
  pp. 2775--2782.

\bibitem{quattoni2009recognizing}
A.~Quattoni and A.~Torralba, ``Recognizing indoor scenes,'' in
  \emph{International Conference on Computer Vision and Pattern
  Recognition}.\hskip 1em plus 0.5em minus 0.4em\relax IEEE, 2009.

\bibitem{razavian2014cnn}
A.~S. Razavian, H.~Azizpour, J.~Sullivan, and S.~Carlsson, ``Cnn features
  off-the-shelf: an astounding baseline for recognition,'' \emph{arXiv preprint
  arXiv:1403.6382}, 2014.

\bibitem{ILSVRCarxiv14}
O.~Russakovsky, J.~Deng, H.~Su, J.~Krause, S.~Satheesh, S.~Ma, Z.~Huang,
  A.~Karpathy, A.~Khosla, M.~Bernstein, A.~C. Berg, and L.~Fei-Fei, ``Imagenet
  large scale visual recognition challenge,'' 2014.

\bibitem{sadeghi2012latent}
F.~Sadeghi and M.~F. Tappen, ``Latent pyramidal regions for recognizing
  scenes,'' in \emph{European Conference on Computer Vision}.\hskip 1em plus
  0.5em minus 0.4em\relax Springer, 2012, pp. 228--241.

\bibitem{shabou2012locality}
A.~Shabou and H.~LeBorgne, ``Locality-constrained and spatially regularized
  coding for scene categorization,'' in \emph{International Conference on
  Computer Vision and Pattern Recognition}.\hskip 1em plus 0.5em minus
  0.4em\relax IEEE, 2012, pp. 3618--3625.

\bibitem{silberman11indoor}
N.~Silberman and R.~Fergus, ``Indoor scene segmentation using a structured
  light sensor,'' in \emph{International Conference on Computer Vision
  Workshops}, 2011.

\bibitem{singh2012unsupervised}
S.~Singh, A.~Gupta, and A.~A. Efros, ``Unsupervised discovery of mid-level
  discriminative patches,'' in \emph{European Conference on Computer
  Vision}.\hskip 1em plus 0.5em minus 0.4em\relax Springer, 2012, pp. 73--86.

\bibitem{sun2013learning}
J.~Sun and J.~Ponce, ``Learning discriminative part detectors for image
  classification and cosegmentation,'' in \emph{International Conference on
  Computer Vision}.\hskip 1em plus 0.5em minus 0.4em\relax IEEE, 2013, pp.
  3400--3407.

\bibitem{tao2013rank}
D.~Tao, L.~Jin, Z.~Yang, and X.~Li, ``Rank preserving sparse learning for
  kinect based scene classification.'' \emph{IEEE transactions on cybernetics},
  vol.~43, no.~5, p. 1406, 2013.

\bibitem{torresani2010efficient}
L.~Torresani, M.~Szummer, and A.~Fitzgibbon, ``Efficient object category
  recognition using classemes,'' in \emph{European Conference on Computer
  Vision}.\hskip 1em plus 0.5em minus 0.4em\relax Springer, 2010, pp. 776--789.

\bibitem{tuytelaars2007vector}
T.~Tuytelaars and C.~Schmid, ``Vector quantizing feature space with a regular
  lattice,'' in \emph{International Conference on Computer Vision}.\hskip 1em
  plus 0.5em minus 0.4em\relax IEEE, 2007, pp. 1--8.

\bibitem{vedaldi08vlfeat}
A.~Vedaldi and B.~Fulkerson, ``{VLFeat}: An open and portable library of
  computer vision algorithms,'' \url{http://www.vlfeat.org/}, 2008.

\bibitem{wang2009evaluation}
H.~Wang, M.~M. Ullah, A.~Klaser, I.~Laptev, C.~Schmid \emph{et~al.},
  ``Evaluation of local spatio-temporal features for action recognition,'' in
  \emph{British Machine Vision Conference}, 2009.

\bibitem{wu2009beyond}
J.~Wu and J.~M. Rehg, ``Beyond the euclidean distance: Creating effective
  visual codebooks using the histogram intersection kernel,'' in
  \emph{International Conference on Computer Vision}.\hskip 1em plus 0.5em
  minus 0.4em\relax IEEE, 2009, pp. 630--637.

\bibitem{wu2011centrist}
------, ``Centrist: A visual descriptor for scene categorization,''
  \emph{Transactions on Pattern Analysis and Machine Intelligence}, vol.~33,
  no.~8, pp. 1489--1501, 2011.

\bibitem{xiao2010sun}
J.~Xiao, J.~Hays, K.~A. Ehinger, A.~Oliva, and A.~Torralba, ``Sun database:
  Large-scale scene recognition from abbey to zoo,'' in \emph{International
  Conference on Computer Vision and Pattern Recognition}.\hskip 1em plus 0.5em
  minus 0.4em\relax IEEE, 2010, pp. 3485--3492.

\bibitem{yao2012describing}
J.~Yao, S.~Fidler, and R.~Urtasun, ``Describing the scene as a whole: Joint
  object detection, scene classification and semantic segmentation,'' in
  \emph{Computer Vision and Pattern Recognition (CVPR), 2012 IEEE Conference
  on}.\hskip 1em plus 0.5em minus 0.4em\relax IEEE, 2012, pp. 702--709.

\bibitem{yuan2011mining}
J.~Yuan, M.~Yang, and Y.~Wu, ``Mining discriminative co-occurrence patterns for
  visual recognition,'' in \emph{International Conference on Computer Vision
  and Pattern Recognition}.\hskip 1em plus 0.5em minus 0.4em\relax IEEE, 2011,
  pp. 2777--2784.

\bibitem{zeiler2014visualizing}
M.~D. Zeiler and R.~Fergus, ``Visualizing and understanding convolutional
  networks,'' in \emph{European Conference on Computer Vision}.\hskip 1em plus
  0.5em minus 0.4em\relax Springer, 2014, pp. 818--833.

\bibitem{zeiler2011adaptive}
M.~D. Zeiler, G.~W. Taylor, and R.~Fergus, ``Adaptive deconvolutional networks
  for mid and high level feature learning,'' in \emph{International Conference
  on Computer Vision}.\hskip 1em plus 0.5em minus 0.4em\relax IEEE, 2011, pp.
  2018--2025.

\bibitem{zheng2012learning}
Y.~Zheng, Y.-G. Jiang, and X.~Xue, ``Learning hybrid part filters for scene
  recognition,'' in \emph{European Conference on Computer Vision}.\hskip 1em
  plus 0.5em minus 0.4em\relax Springer, 2012, pp. 172--185.

\bibitem{zhu2010large}
J.~Zhu, L.-J. Li, L.~Fei-Fei, and E.~P. Xing, ``Large margin learning of
  upstream scene understanding models,'' in \emph{Advances in Neural
  Information Processing Systems}, 2010, pp. 2586--2594.

\end{thebibliography}
\end{document}